\documentclass[letterpaper, 10 pt, conference]{ieeeconf}  % Comment this line out if you need a4paper

\IEEEoverridecommandlockouts                              % This command is only needed if 
\usepackage{graphics} % for pdf, bitmapped graphics files
\usepackage{epsfig} % for postscript graphics files
\usepackage{mathptmx} % assumes new font selection scheme installed
\usepackage{times} % assumes new font selection scheme installed
\usepackage{amsmath} % assumes amsmath package installed
\usepackage{amssymb}  % assumes amsmath package installed
\usepackage[noadjust]{cite}

\usepackage{comment}

\usepackage[pagebackref,breaklinks,colorlinks,citecolor=cvprblue]{hyperref}
\usepackage{booktabs}
\usepackage{pifont}% http://ctan.org/pkg/pifont
\usepackage{algorithm}
\usepackage{algpseudocode}
\usepackage{graphicx}
\usepackage{float}
\usepackage{tabularx,colortbl}
\usepackage{xparse,mathtools}
\usepackage{diagbox}
\usepackage{adjustbox,lipsum}
\usepackage{breqn}

\mathchardef\mhyphen="2D
\usepackage[utf8]{inputenc}

\usepackage{tabulary,overpic}

\newlength\savewidth

\title{\LARGE \bf
Scalable Offline Metrics for Autonomous Driving
}

\usepackage[capitalize]{cleveref}
\crefname{section}{Sec.}{Secs.}
\Crefname{section}{Section}{Sections}
\Crefname{table}{Table}{Tables}
\crefname{table}{Tab.}{Tabs.}

\usepackage{bbm}
\usepackage{xspace}
\usepackage{dsfont}
\usepackage{multirow}
\newcommand{\ba}{\mathbf{a}}

\newcommand{\bI}{\mathbf{I}}

\newcommand{\bL}{\mathbf{L}}

\newcommand{\bo}{\mathbf{o}}

\newcommand{\btheta}{\boldsymbol{\theta}}

\newcommand{\cA}{\mathcal{A}}

\newcommand{\cL}{\mathcal{L}}

\newcommand{\cO}{\mathcal{O}}

\makeatletter
\DeclareRobustCommand\onedot{\futurelet\@let@token\@onedot}
\def\@onedot{\ifx\@let@token.\else.\null\fi\xspace}
\def\eg{e.g\onedot} 
\def\ie{i.e\onedot}

\def\wrt{wrt\onedot}

\def\etal{et~al\onedot}

\makeatother

\newcommand{\boldparagraph}[1]{\vspace{0.2cm}\noindent{\bf #1:}}

\definecolor{darkgreen}{rgb}{0,0.7,0}

\definecolor{lightgray}{rgb}{0.89, 0.89, 0.89}

\author{Animikh Aich*, Adwait Kulkarni*, and Eshed Ohn-Bar$^{1}$% <-this % stops a space
\thanks{$^{1}$Authors are affiliated with the College of Engineering, Boston University, Boston, MA 02215, USA 
        {\tt\small \{animikh, adk1361, eohnbar\}@bu.edu}. 
        \newline*These authors contributed equally to this work.}%
}

\begin{document}

\maketitle
\thispagestyle{empty}
\pagestyle{empty}

\begin{abstract}

Real-world evaluation of perception-based planning models for robotic systems, such as autonomous vehicles, can be safely and inexpensively conducted offline, \ie by computing model prediction error over a pre-collected validation dataset with ground-truth annotations. However, extrapolating from offline model performance to online settings remains a challenge. In these settings, seemingly minor errors can compound and result in test-time infractions or collisions. This relationship is understudied, particularly across diverse closed-loop metrics and complex urban maneuvers. In this work, we revisit this undervalued question in policy evaluation through an extensive set of experiments across diverse conditions and metrics. Based on analysis in simulation, we find an even worse correlation between offline and online settings than reported by prior studies, casting doubts on the validity of current evaluation practices and metrics for driving policies. Next, we bridge the gap between offline and online evaluation. We investigate an offline metric based on \textit{epistemic uncertainty}, which aims to capture events that are likely to cause errors in closed-loop settings. The resulting metric achieves over 13\% improvement in correlation compared to previous offline metrics. We further validate the generalization of our findings beyond the simulation environment in real-world settings, where even greater gains are observed.
\end{abstract}

%test\cite
\section{Introduction}
\label{sec:intro}
Can we estimate the online performance of a driving model without actively driving with it? Evaluating driving policy models that map camera input to a driving plan or controller action in real-world conditions is typically expensive and potentially hazardous.
During model deployment, traffic infractions and collisions may occur, such as fatal accidents with pedestrians and vehicles \cite{uber,cruiseend}, Waymo's driving in the wrong lane and veering unsafely \cite{waymodrunk}, and Tesla's Autopilot crashing into a truck on a busy highway \cite{ghorai2024causation}.   

\begin{figure}[!t]
    \centering
    \small
    \begin{tabular}{c}
        \includegraphics[width=1\columnwidth, trim=0 40mm 80mm 0mm, clip]{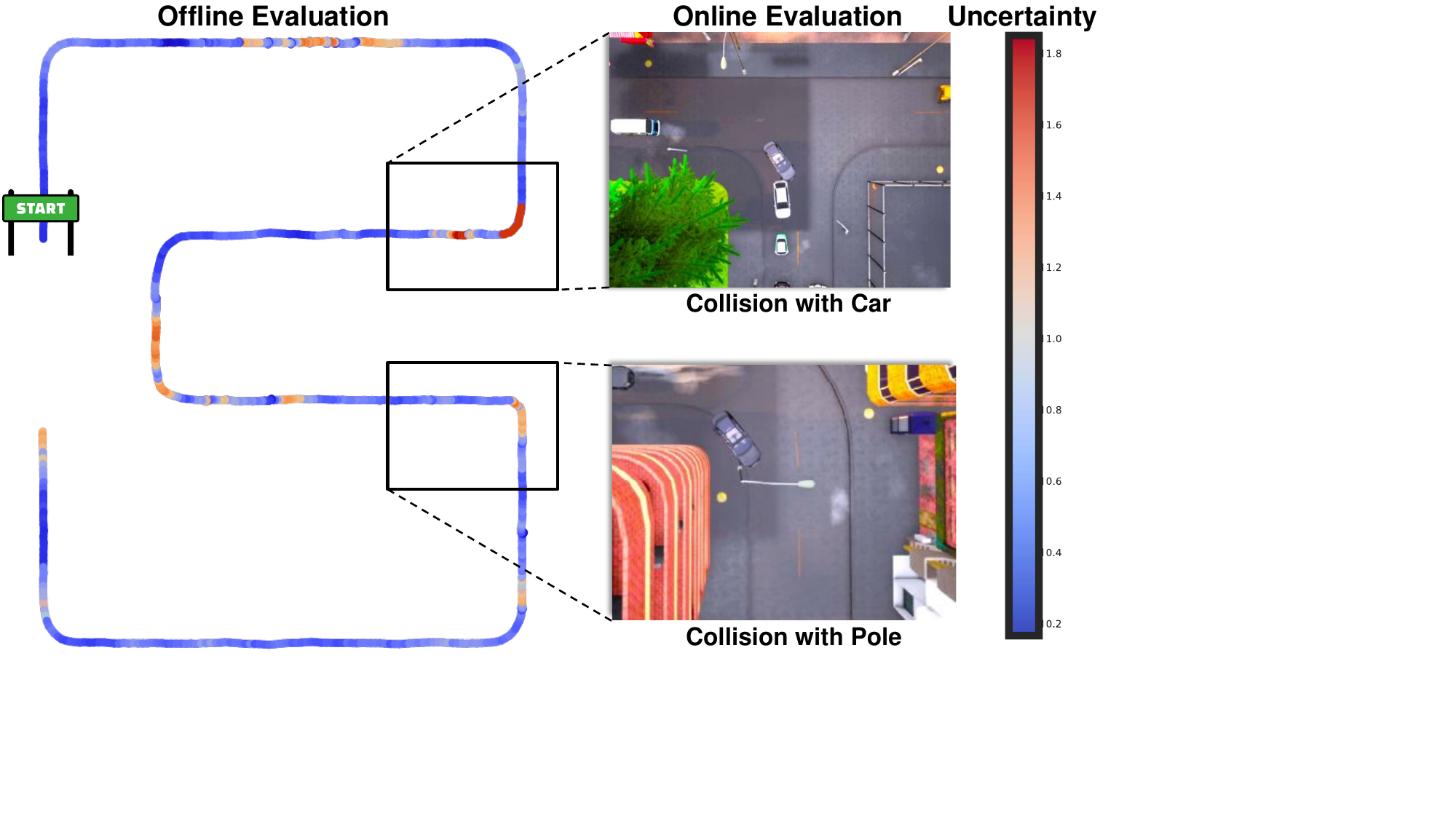}
    \end{tabular}
    \vspace{-0.1cm}
    \caption{\textbf{Closing the Gap Between Offline and Online Evaluation for Autonomous Driving.} We revisit the relationship between offline and online evaluation of vision-based autonomous driving policies. Our key insight is that model uncertainty-weighted errors (shown on the left) can be used to estimate errors in online settings (shown on the right), such as collisions and traffic infractions. We then devise a simple and scalable metric that can be applied over offline data without requiring high-quality perception information or various surrounding agent prediction models. We validate the effectiveness of the metrics in both simulated and real-world settings across diverse model sampling strategies, types of policy failures, and platforms. 
    }
    \label{fig:teaser}
\end{figure}

To circumvent such safety-critical evaluations, researchers often use an offline approach. They leverage a computed \textit{expected loss} over a pre-collected, real-world dataset. This method avoids the potentially dangerous and costly quantification of collisions and traffic infractions in live scenarios (Fig.~\ref{fig:teaser}). The efficiency in offline evaluation over a pre-collected driving dataset, where the driving model predictions are compared against safe ground-truth targets provided by a human driver (\eg, leveraging an L2 loss~\cite{hu2021safe,behl2020label,hu2023planning,zhang2023coaching,jaeger2023hidden,jiang2023vad,wang2019monocular,Codevilla2018ICRA,zhu2023learning}), has recently propelled significant advancements in scalable autonomous driving models~\cite{bansal2018chauffeurnet, hu2023planning,jiang2023vad,wu2023policy,dauner2023parting,hu2022st,zhang2024feedback,zhang2022rethinking}. However, it remains unclear whether current offline evaluation practices adequately reflect actual on-road driving model performance, particularly for decisions that require highly robust and split-second reactions with life-threatening implications.

The reasons for the discrepancy between offline and online model performance are three-fold. First, real-world driving datasets are nearly entirely \textit{uneventful}, with drivers rarely experiencing safety-critical events such as abrupt pedestrian crossings, near-collisions, or even turns and merging. The expected loss over the inherently imbalanced distribution of data can obscure performance under safety-critical events~\cite{tu2023towards,li2023ego}. Second, while automatic data sampling techniques can alleviate data imbalance issues~\cite{thorn2018framework,menzel2018scenarios,o2018scalable,weber2019framework,lee2020adaptive}, the standard L2 loss-based metric itself provides a crude measure for the ultimate task of safe and enjoyable driving~\cite{li2020end,Codevilla2018ECCV}. For instance, the L2 metric can remain identical among several maneuvers, \eg, merging to the left or right lane on a freeway, yet one decision may be perfectly safe while the other results in a collision, \ie, in the case of an adjacent vehicle in a nearby lane. This observation holds for nearly any scenario in complex urban driving settings, as shown in Fig.~\ref{fig:teaser}. Additional open-loop metrics have been devised, such as those based on future collision estimates~\cite{hu2021safe,hu2023planning,jaeger2023hidden,jiang2023vad,li2023ego}. However, these metrics induce similar evaluation issues. Their reliance on various naive velocity models prevents them from capturing realistic interactive navigation and negotiation in intricate and dense settings. Moreover, collision metrics lack scalability due to the requirement of perception annotations~\cite{schreier2023offline}. Third, online model testing is done in a \textit{closed-loop manner}, where minor decision errors can propagate and result in unpredictable model behavior over time in testing~\cite{ross2011reduction,osa2018algorithmic,prakash2020exploring,laskey2017dart}. 

Despite known limitations in offline open-loop evaluation~\cite{zhang2022rethinking,li2023ego}, prior work has rarely studied extrapolation from offline to online settings beyond simplistic settings. For instance, Codevilla~\etal~\cite{Codevilla2018ECCV} demonstrated the limited correlation between the two settings but only in the simplest towns and maneuvers in the CARLA simulation~\cite{Dosovitskiy2017CORL}, \ie, Town01 and Town02, with one online metric (success rate), and partial action values (only steer, excluding throttle and brake ground truth). Therefore, generalization to intricate maneuvers and complex urban scenarios, diverse online metrics, or real-world settings remains understudied despite its broad implications. In this work, we revisit the question of the correlation between offline and online evaluations. 

\boldparagraph{Contributions} We aim to advance safe and reliable evaluation procedures for autonomous driving policies. Our key insights are threefold. \textit{First}, we leverage a more expansive evaluation in simulation, covering multiple towns and closed-loops metrics. Our results reveal even worse correlation patterns than have been previously shown. \textit{Subsequently}, we hypothesize that the poor correlation is due to degraded model performance during instances of uncertainty, \ie, challenging or rare events in the data. We thus analyze an effective metric based on an \textit{uncertainty-weighted error}, and demonstrate its broad effectiveness across settings, models, and error types. While uncertainty-based techniques are standard in robotics and machine learning~\cite{filos2020can,stachowicz2024racer,ovadia2019can}, to the best of our knowledge, we are the first to quantify their role in estimating online evaluation of driving policies.  We observe a significant improvement, \eg, 13\% higher correlation in CARLA. \textit{Third}, given challenges in realistic simulation of real-world physics and appearance, we go beyond simplistic simulations to characterize correlation trends in the real world using a small-scale platform, further validating our proposed metric.
\section{Related Work}
\label{sec:related}

\boldparagraph{End-to-End Autonomous Driving}
With the emergence of deep learning models, autonomous driving research has undergone a significant transformation, shifting from rule-based systems to learning-based approaches. This shift is notably reflected in the proliferation of end-to-end autonomous driving policies utilizing deep learning techniques~\cite{bojarski2016end,lai2024uncertainty, xu2017end, chitta2022transfuser, zhang2023coaching, pomerleau1989alvinn, bansal2018chauffeurnet, Cui_2021_ICCV,behl2020label,ohn2020learning, hu2022st}. This trend is further exemplified by the CARLA Leaderboard~\cite{carlaleaderboard}, where deep learning-based methods have gained prominence, surpassing modular approaches in autonomous driving performance~\cite{muller2018driving, tang2018teaching, chen2021what, gonzalez2015review, kendall2019learning, xu2021autonomous}. 
In addition to camera-based RGB images, additional sensors, such as LiDAR, have also been shown to play a key role in improving the safety of driving policies through multi-modal sensor fusion~\cite{chitta2022transfuser, shao2023safety, prakash2021multi}.

\boldparagraph{Issues with Offline Evaluation}
The progress of learning-based models is primarily driven by offline evaluation methods. However, the question of their accuracy in assessing driving performance within online settings still remains. Codevilla et al.~\cite{Codevilla2018ECCV} is one of the first to attempt to study this problem and proposes several novel offline evaluation metrics, including speed-weighted and quantized versions of standard loss metrics. Among the metrics, the NuScenes Detection Score (NDS)~\cite{caesar2020nuscenes} is a widely accepted offline metric that has been shown in a recent work by Schreier et al.~\cite{schreier2023offline} to provide improved correlation with online metrics like CARLA Driving Score~\cite{Dosovitskiy2017CORL}. As shown by Li et al. in \cite{li2023ego}, a simple baseline model without vision can achieve competitive status on a leaderboard evaluated by offline metrics like NDS, thus proving the limitations of existing offline metrics in predicting driving performance. These recent discussions demand a revisit on whether offline metrics are sufficient in predicting driving performance.

\boldparagraph{Safety-Critical Scenario Analysis}
Safety critical scenarios refer to specific scenarios, such as collisions, where safety is compromised, and the ego vehicle must take preventative or evasive actions to avoid catastrophic outcomes~\cite{ghorai2024causation,wang2021advsim,weber2019framework,hanselmann2022king}. Current methodologies address this challenge by curating datasets with generated safety-critical scenarios~\cite{lee2020adaptive,lai2023xvo,menzel2018scenarios,thorn2018framework}. This approach allows for testing on various generated scenarios to validate if the policy is safe. While we share a similar goal, this is an orthogonal direction to our approach as we focus on designing a useful metric across datasets. However, while generated scenario realism and coverage remain an open question, our proposed metric can be automatically and easily applied at scale to any policy and dataset (\ie, only based on the model and without requiring privileged annotated information).

\boldparagraph{Off-Policy Evaluation}
Offline Policy Evaluation (OPE) in RL~\cite{review-off-RL-tutorial}, as well as offline RL~\cite{off-RL-tutorial}, is a related line of research, aiming to minimize the need for direct interaction with the environment to train or evaluate an RL agent. We use an instantiation of OPE and interpret it as a sampling mechanism~\cite{instantiation-uncertainty-based-metric} in our proposed methods. OPE approaches are also known to suffer from high variance in high-dimensional and complex settings. In general, RL and offline RL-based methods have yet to be adapted to complex environments such as CARLA, i.e., moving beyond tabular or simple settings often studied in OPE.

\boldparagraph{Non-Reactive Simulation Benchmarks}
Recent work by Dauner et al.~\cite{dauner2024navsim} introduces NAVSIM, a data-driven non-reactive simulation framework that bridges offline and online evaluation by unrolling short-horizon simulations on real-world sensor data. It computes safety-critical metrics such as collisions and progress, using bird’s-eye-view abstractions. This approach has been shown to outperform traditional displacement errors (e.g., ADE) in closed-loop alignments. While NAVSIM enables scalable benchmarking (e.g., revealing lightweight models like TransFuser~\cite{chitta2022transfuser} can match complex architectures), its non-reactive assumption limits modeling of compounding errors or interactive scenarios. Our work complements NAVSIM by proposing uncertainty-weighted metrics that address rare or uncertain failures without requiring perception annotations, improving scalability in reactive or long-horizon settings.
% \vspace{-0.02cm}

\section{Method}
\label{sec:method}
\begin{table*}[!t]
\centering
\caption{\textbf{Summary of Offline and Online Metrics.} An overview of metric terms. The list is not exhaustive but compares the most commonly employed online metrics and the baseline offline metrics.}
\label{tab:metrics}
\begin{tabular}{>{\raggedright\arraybackslash}p{0.3\linewidth} >{\raggedright\arraybackslash}p{0.6\linewidth}}
\toprule
 \rowcolor{lightgray}
\multicolumn{2}{c}{\textbf{Offline}} \\ \midrule
\textbf{Metric} & \textbf{Definition} \\ \midrule
Steer MSE & MSE of predicted steering and expert steering. \\ \hline
SW Steer MSE & Speed weighted MSE of predicted steering angle and expert steering angle.  \\ \hline
UW Steer MSE & Uncertainty weighted MSE of predicted steering angle and expert steering angle. \\ \hline
Steer MAE & MAE of predicted steering angle and expert steering angle. \\ \hline
SW Steer MAE & Speed weighted MAE of predicted steering angle and expert steering angle. \\ \hline
UW Steer MAE & Uncertainty weighted MAE of predicted steering angle and expert steering angle. \\ \hline
Throttle MAE & MAE of predicted throttle value and expert throttle value. \\ \hline
Waypoint MAE & MAE of predicted waypoints and expert waypoints. \\ \hline
Waypoint FDE & Final displacement error of final predicted waypoints and expert final waypoints. \\ \hline
TRE & Thresholded relative error. \\ \hline
QCE & Quantized classification error. \\ \hline
FDE & Error between final predicted displacement and expert final displacement. \\ \hline
Action MSE & MSE of combined throttle and steering. \\ \hline
Action MAE & MAE of combined throttle and steering. \\ \hline
UW Action MSE & Uncertainty weighted MSE of combined throttle and steering. \\ \hline
UW Action MAE & Uncertainty weighted MAE of combined throttle and steering. \\ \hline
PDM Score & Composite metric evaluating collisions, progress, and comfort in non-reactive simulations. \\ \midrule

\rowcolor{lightgray}
\multicolumn{2}{c}{\textbf{Online}} \\ \midrule
\textbf{Metric} & \textbf{Definition} \\ \midrule
Outside Route Lane & The number of times the wheels of the car cross the lane of the assigned route. \\ \hline
Route Deviation & More than 30-meter deviation from route assigned. \\ \hline
Route Timeout & Takes too long to complete route assigned. \\ \hline
Vehicle Blocked & Vehicle does not take an action for more than 180 seconds. \\ \hline
Driving Score & Product of route completion and the infractions penalty. \\ \hline
Success & Vehicle completes the assigned route, denoted by 0 (incomplete) or 1 (complete). \\ \hline
Route Completion & Percentage of the route distance completed by vehicle. \\ \hline
Infractions & Total number of infractions occurred in one route. \\ \hline
Collisions (All) & Number of times vehicle did not stop when it was supposed to. \\ \hline
Collisions (Vehicle) & Number of times the vehicle collided with another vehicle. \\ \hline
Collisions (Environment) & Number of times the vehicle collided with its surroundings. \\ \hline
Red Light Violation & Number of times the vehicle did not stop in the presence of a red light. \\ \hline
Stop Infraction & Number of times the vehicle did not stop in the presence of a stop sign. \\      \bottomrule
    \end{tabular}
    % \vspace{-0.1cm}
\end{table*}

Our goal is to estimate online performance through offline evaluation, \ie, errors computed over a dataset of safely demonstrated driving trajectories. We first formulate our task in Sec.~\ref{subsec:data}. Then, we define a comprehensive set of offline and online metrics that measure different aspects of model predictions in Sec.~\ref{subsec:metrics}. 

\subsection{Settings and Data}
\label{subsec:data}
% \vspace{-0.5cm}
To validate model performance, we follow standard protocols in vision-based, end-to-end motion planners and open-source simulations. 
In our settings, we assume access to a collected dataset $\mathcal{D} = \{(\bo_i, \ba_i)\}_{i=1}^N$. Each observation $\bo_i = (\bI_i, \bL_i, c_i, v_i) \in \cO$, comprises an image $\bI_i \in \mathbb{R}^{N_W \times N_H \times 3}$, a LiDAR point cloud with alpha channel $\bL_i \in \mathbb{R}^{N_X \times N_Y \times N_Z\times 1}$, conditional command $c_i \in \mathbb{N}$, and speed measurements $v_i \in \mathbb{R}$, and ground-truth action vector $\ba_i \in \cA$ executed by the driver. We denote $\pi_{\btheta}:\cO\to\cA$ as the policy, a mapping function, parameterized using weights $\btheta \in \mathbb{R}^d$, that predicts actions $\hat{\ba}_i$ based on input observations $\bo_i$.
%\mathbb{R}^{N_a}
We note that the action may either be a low-level control action, \eg, a 3D vector of steering, throttle, and brake amounts, or a set of future 2D waypoints in the map's view representing an intended trajectory to follow by the vehicle~\cite{Codevilla2018ECCV,muller2018driving}. Vision-based planners today generally employ waypoints as the model output~\cite{jiang2023vad,hu2023planning,chitta2022transfuser}, where a low-level controller (\eg, PID) can be used to produce the final control. In contrast, prior work emphasizes direct steer value prediction~\cite{Codevilla2018ECCV}. Given the widespread use of waypoints-based metrics (defined below), our analysis explores the role of both action definitions and their correlation with online, closed-loop performance.

\begin{comment}
    
\end{comment}

% \vspace{-0.2cm}
\subsection{Metrics}
\label{subsec:metrics}
% \vspace{-0.1cm}
Given a loss $\cL$ and a per-sample scalar weight $\alpha_i \in \mathbb{R}$, an offline metric can be defined as an expectation \wrt a dataset:
\begin{equation}
\label{eqn:opt}
% \argmin_{\pi} \mathbb{E}_{(\mathcal{X}, \mathcal{W}) \sim \mathcal{D}} [\mathcal{L}(\mathcal{W}, \pi(\mathcal{X}))]
%\cE = 
\mathbb{E}_{(\bo_i, \ba_i) \sim \mathcal{D}} [\alpha_i  \mathcal{L}(\ba_i, \pi_{\btheta} (\bo_i)) ]
\end{equation}
A choice of setting weight scalars to $\alpha_i=1$ and setting an L1 loss, \ie, $\cL = \|{\mathbf{a}_i - \mathbf{\hat{a}}_i}\|_1$, gives the mean absolute error (MAE) metric. Similarly, mean squared error (MSE) is given by using the L2 loss~\cite{hu2023planning,li2020end}. In principle, a large number of different offline metrics may be derived for a given dataset, yet only a few have been shown to produce a correlation with online performance, particularly for policy's overall success rate along a route known as the \textit{QCE} and \textit{TRE} (based on Codevilla~\etal~\cite{Codevilla2018ECCV}), defined next for clarity. For clarity, we summarize all studied metrics in Table~\ref{tab:metrics}.

\begin{comment}

\end{comment}

%small errors are ignored and larger errors are identified.
\boldparagraph{Quantized Classification Error (QCE)} 
QCE quantizes the predictions such that only larger errors from the ground truth are considered, defined via a threshold. QCE is given by:
\begin{equation}
\label{eqn:qce}
%     \cL_{QCE} = \frac{1}{|{V}|}\sum\limits_{i \in V} (1 - \delta(Q({\ba}_i, \sigma), Q(\hat{\ba}_i, \sigma)))
    \cL_{QCE} = (1 - \delta(Q({\ba}_i, \sigma), Q(\hat{\ba}_i, \sigma)))
\end{equation}
Where $\delta$ is the Kronecker delta function and $Q(x)$ is given by:
\begin{equation}
    Q(x) = \begin{cases} -1 & \text{if } x < -\sigma \\ 0 & \text{if } -\sigma \le x < \sigma \\ 1 & \text{if } x > \sigma \end{cases}
\end{equation}
To ensure meaningful comparison, we employ the same threshold settings provided in Codevilla~\etal~\cite{Codevilla2018ECCV}.

\boldparagraph{Thresholded Relative Error (TRE)} TRE builds on QCE by making use of an adaptive threshold proportional to the ground truth steering angle. It penalizes errors during small ground truth action values, \ie, when the steering angle is low. TRE is given by: 
\begin{equation}
\label{eqn:tre}
    % \cL_{TRE} = \frac{1}{|{V}|}\sum\limits_{i \in V} H(\|\mathbf{\hat{a}}_i - \mathbf{a}_i\| - \alpha \|{\mathbf{a}_i}\|)
    \cL_{TRE} = H(\|\mathbf{\hat{a}}_i - \mathbf{a}_i\| - \lambda\|{\mathbf{a}_i}\|)
\end{equation}
where $H$ is the Heaviside step function. We set $\lambda = 0.1$ consistently with \cite{Codevilla2018ECCV} and $\sigma = 0.5$ as the midpoint of our steering range. We note that we set $\alpha_i=1$ in general for computing errors such as MSE, QCE, and TRE. However, we hypothesize that online errors occur in scenarios where the model may be uncertain about its outputs. We thus propose to leverage the weights $\alpha_i$ to proportionally weigh such instances, as defined next.

\boldparagraph{Uncertainty-Weighted Error (UWE)} Codevilla~\etal~\cite{Codevilla2018ECCV} were motivated by the intuition that errors at higher ego speeds have a greater impact on driving performance. They studied a speed-weighted metric where the weights in Eq.~\ref{eqn:opt} are set to the speed, \ie, $\alpha_i = v_i$. However, this did not result in a better correlation with the route's success rate. Instead, we propose to leverage a model-dependent, \textit{uncertainty-based weight}. We employ monte carlo (MC) dropout \cite{gal2016dropout} to efficiently estimate epistemic uncertainty. This involves enabling dropout during the prediction phase and performing $K$ forward passes through the network for each input sample $i$. We then calculate the variance $u_i$ of these predictions to estimate the model's uncertainty~\cite{lai2025zerovo}.
Then, $u_i$ is raised to $\gamma$ to scale the variance's importance, and a weighted sum is applied to various offline metrics, each with optimized weights $\beta_t$ to compute the UWE:
\begin{multline}
\label{eqn:uw}
    \text{UWE} = \sum_{t=1}^{n} \beta_t \mathbb{E}_{(\bo_i, \ba_i) \sim \mathcal{D}} [u_i^\gamma  \mathcal{L}_t(\ba_i, \pi_{\btheta}, (\bo_i)) ], s.t. \\ 
    u_i = \text{var}\left(\{ \pi_{\btheta_k}(\bo_i)  \}_{k=1}^K \right)
\end{multline}

where $n$ is the number of offline metrics and $\mathcal{L}_t$ is different offline metric. A weighted sum of the different uncertainty-based weighted offline metrics is able to capture various edge cases from each offline metric. While uncertainty has been applied to improve and regularize a range of robot perception and learning tasks~\cite{kendall2018multi,ovadia2019can,henaff2019model,wang2021uncertainty}, we are the first to measure its utility in the context of offline and online policy evaluation. The MC Dropout-based uncertainty provides a simple and efficient weight that can be applied to any prior offline metric, yet we find its combinations with the basic errors, \ie, MAE and MSE, sufficient.

\boldparagraph{Online Metrics} Online metrics convey information about the driving performance in on-policy test time settings. We follow metrics defined in CARLA Leaderboard \cite{carlaleaderboard,chitta2022transfuser}, including the rate of various traffic incidences, such as Collisions and Stop Infractions. The main metric in CARLA is the Driving Score (DS). DS is the average of the product of the Route Completion (\% from the planned route) score $R_i$ and an Infraction penalty $P_i$ across $N$ routes:
\begin{equation}
\label{eq:ip}
   \text{DS}=\frac{1}{N}\sum_{i=1}^{N}R_i P_i
\end{equation}
We emphasize that prior work has only looked into Success Rate (SR) as the main online metric~\cite{Codevilla2018ECCV}. However, SR does not consider route completion (\eg, $95\%$ completion score still means zero success) nor infraction types along the route. Instead, DS provides a holistic evaluation of driving behavior, promoting safer and more efficient driving practices. In this paper, we will study the correlation of offline metrics with various online metrics, with a particular focus on the DS. 

\section{Experiments}
\label{sec:analysis}
We first describe our experimental setup and model training and testing procedure. We then leverage the setup to perform comprehensive experiments in simulation over various towns, ambient settings, and traffic infractions using the CARLA simulation~\cite{Dosovitskiy2017CORL}. However, while most online performance studies for autonomous driving are performed in simulation, this may bias results, \ie, due to appearance and physics artifacts. 
To ensure our findings generalize to real-world models, we thus also incorporate analysis using a small-scale platform in the real-world.

\subsection{Experimental Setup}
\label{subsec:setup}
\boldparagraph{Simulation Benchmark} We follow the standard Longest6 testing setup ~\cite{chitta2022transfuser,jaeger2023hidden}. Longest6 benchmark comprises 36 routes with an average route length of 1.5km, which goes beyond the prior study on CARLA's Town02 by Codevilla~\cite{Codevilla2018ECCV}. We leverage a privileged rule-based expert to generate a large training and validation dataset, where trained policies are evaluated over unseen weather and routes in training.

\begin{comment}

\end{comment}

\begin{figure}[!t]
    \centering
    \vspace{-0.6cm}
    \includegraphics[width=1\linewidth, trim=0 40 0 40, clip]{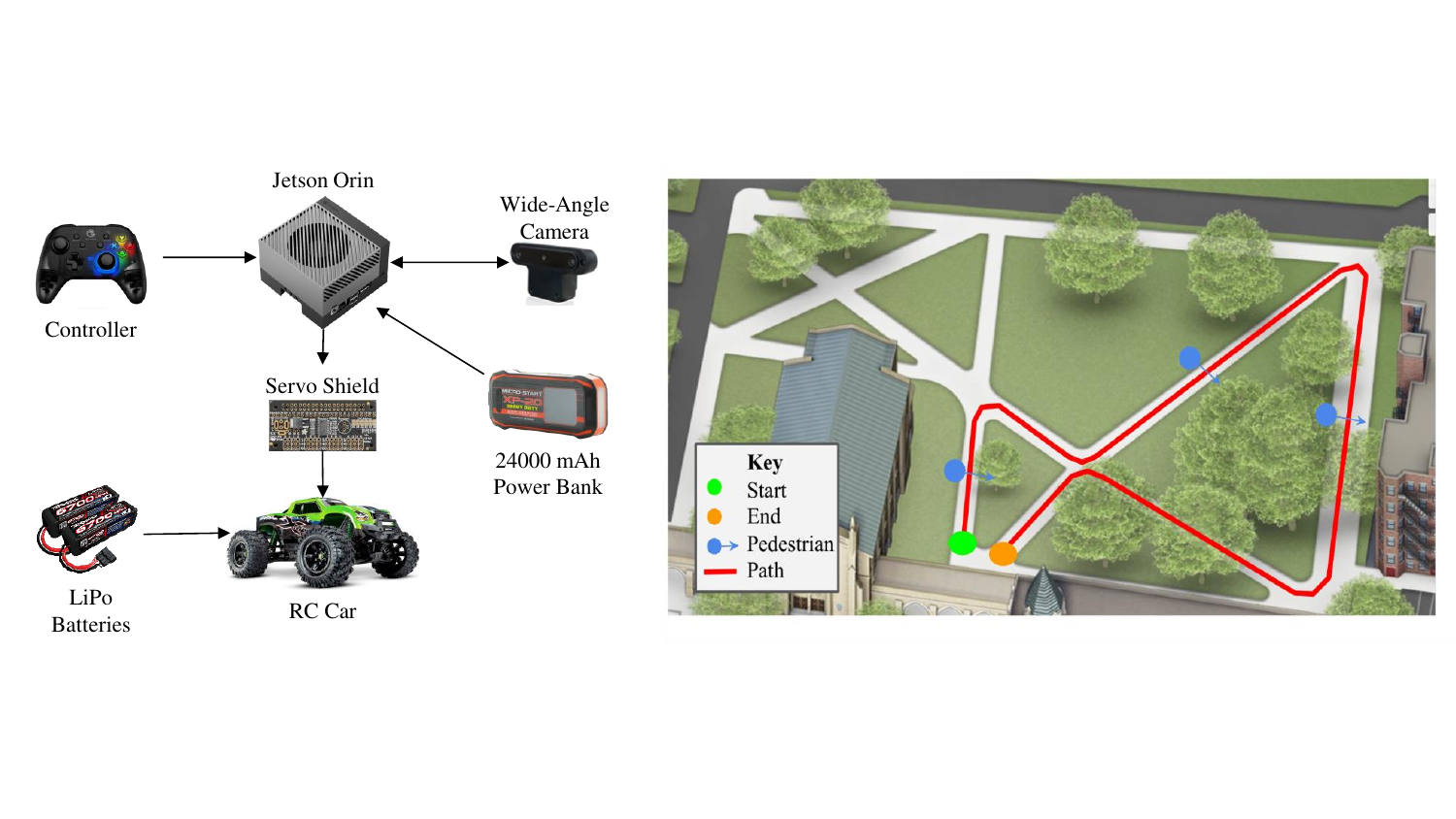}
    \vspace{-1.2cm}
    \caption{\textbf{Real-World Vehicle Platform (Left) and Bird's Eye View of an Example Evaluation Route (Right).} The vehicle is a hacked off-the-shelf RC car (Traxxas XMaxx). It is equipped with a Jetson Orin with camera and is controlled by a joystick controller for data collection and evaluation. }
    %The diagram shows the power and data control paths within the platform. }}
    \label{fig:rw-setup}
    \vspace{-0.5cm}
\end{figure}
\boldparagraph{Real-World Benchmark}
Online evaluation of driving policies is generally done in simulation~\cite{chitta2022transfuser,lbc,jiang2023vad}. To ensure our findings generalize to the real-world, we utilize a safe, small-scale platform. This platform is based on a commercial RC car (Traxxas X-Maxx~\cite{xmax}). We set up a Jetson Orin and a wide-angle camera to capture observations and control the steering and speed of the car using custom scripts and a joystick controller. We note that we do not leverage LiDAR in our real-world platform. We collect a training set of $32,356$ frames on various sidewalks, times of day, and weather. 
We evaluate on an uneven road that has three right turns, two left turns, and three pedestrians crossing (example routes are shown in Fig. \ref{fig:rw-setup}). The platform can travel at speeds of up to $50$MPH. Online evaluations are safely done with an emergency software stop. In addition to the long route evaluation, we also perform targeted evaluations with clearly defined scenarios over traffic lights, vehicles, and object crossings. 

%, which are then inputted to a PID controller
\boldparagraph{Model Training Strategies}
In the simulation, our baseline model is TransFuser~\cite{jaeger2023hidden}, which provides a state-of-the-art conditional imitation learning model. The model is a multi-modal policy with two inputs: RGB image and LiDAR. The features of these modalities are fused together by a transformer-based architecture with a cross-modal attention mechanism to predict output actions. To ensure our findings are applicable to a broad range of models, we train a large set of model variations by modifying the model input (\eg, RGB only vs. RGB and LiDAR) and backbone (two ConvNext~\cite{liu2022convnet}, small and tiny, and nine RegNet~\cite{radosavovic2020designing}, from RegNetY-200MF to RegNetY-8.0GF). We train each model (input and backbone combination) for 40 epochs while also saving the checkpoints every 10 epochs to sample additional model variations. Applying test-time dropouts provides another means of sampling models and covering the space of all possible policies. Moreover, we sample models using test-time dropout for values between 10\% to 30\% with increments of 10\%. The Pearson correlation coefficient is then computed over all sampled models. {Results are presented across 82 models, each varying in backbones, epochs, inputs, and test-time dropout rates}. For real-world models, in the interest of optimizing for real-time performance, we target smaller and simpler architectures motivated by prior works from Codevilla et al.~\cite{Codevilla2018ICRA}, Hawke et al.~\cite{hawke2020urban} and Bojarski et al.~\cite{bojarski2016end} with a baseline of a five-layer CNN as well as variants of RegNetY as feature extractor backbones. In the real-world experiment, we train an end-to-end policy model solely using RGB input, as it provides an easy and scalable implementation for researchers to reproduce and build upon in the future (\ie, without LiDAR).

\begin{figure*}[!t]
    \centering
    \includegraphics[width=1\linewidth]{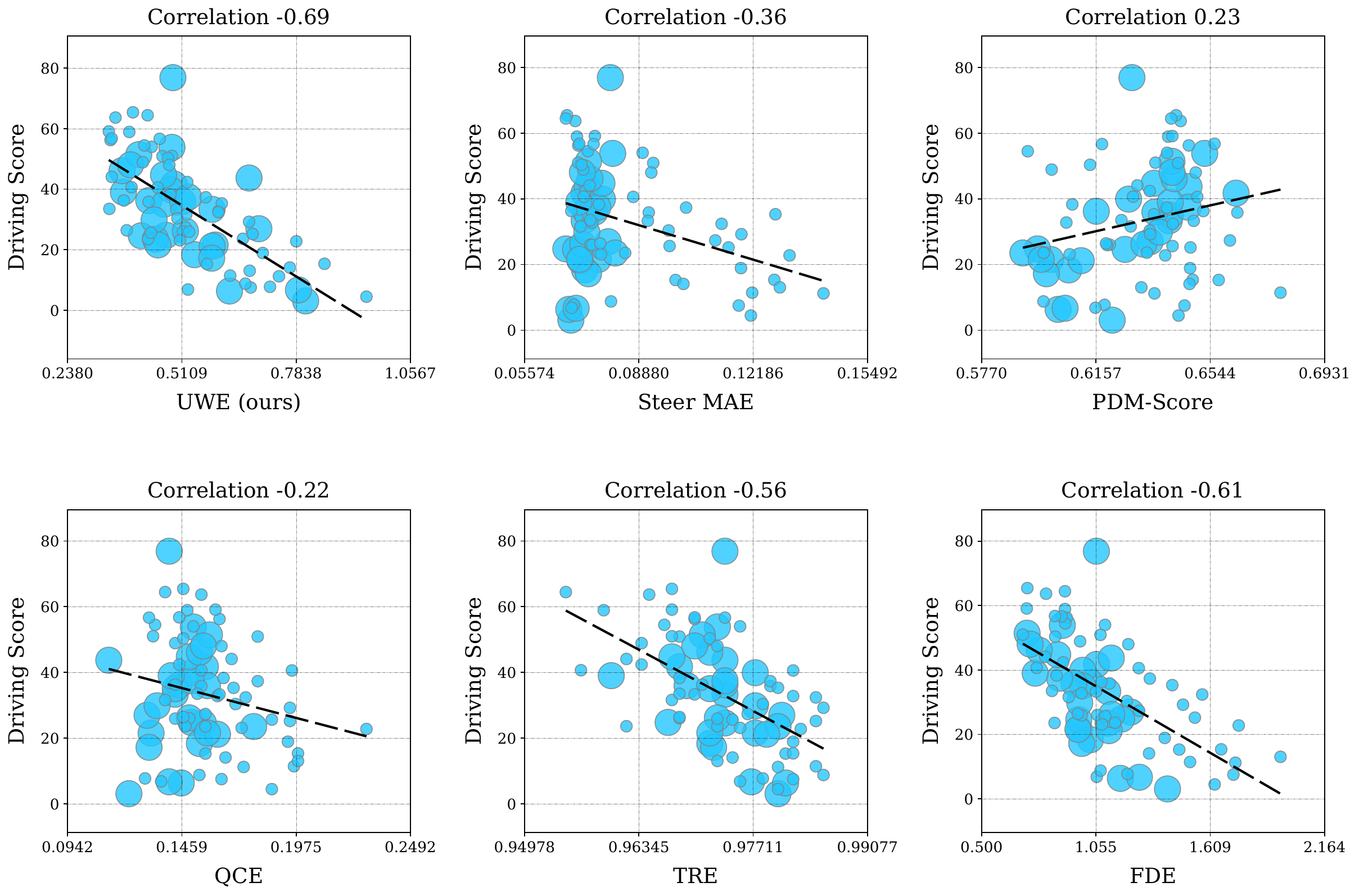}
    \vspace{-0.6cm}
    \caption{{\textbf{Driving Score Correlation Analysis in Simulation.} The plot shows updated correlation for offline metrics, including TRE, QCE and PDM, the most successful reported metrics by prior research~\cite{Codevilla2018ECCV,dauner2024navsim}, as well as widely employed metrics such as waypoint FDE (Final waypoint Displacement Error). Given our updated evaluation setup with complex traffic scenarios and diverse models, correlations are low overall, besides the introduced uncertainty-weighted (UW) version. Each disc shows one model sampled from a certain epoch, backbone, input (with or without LiDAR), and a test-time dropout rate used to obtain coverage of models with varying performances (see further explanation in Sec.~\ref{subsec:setup}). The radius of each marker is proportional to the percent of test-time dropout used (this sampling is independent of our proposed metric).}}
    \label{fig:ds-2x3}
    \vspace{-0.3cm}
\end{figure*}

\subsection{Results in Simulation}

Given the more extensive benchmark and model sampling strategies, \ie, with dropout over model weights for broader coverage of the performance spectrum, we revisit the analysis of Codevilla et al.~\cite{Codevilla2018ECCV} (only analyzed steering-based metrics, while \textit{fixing other action values based on the ground-truth}). We also leverage the Driving Score, as it provides a holistic measure for driving quality, \ie, beyond Success Rate, as well as finer-grained online metrics. 

%such as 

\begin{figure}[htbp]
    \centering
    \includegraphics[width=3.2in]{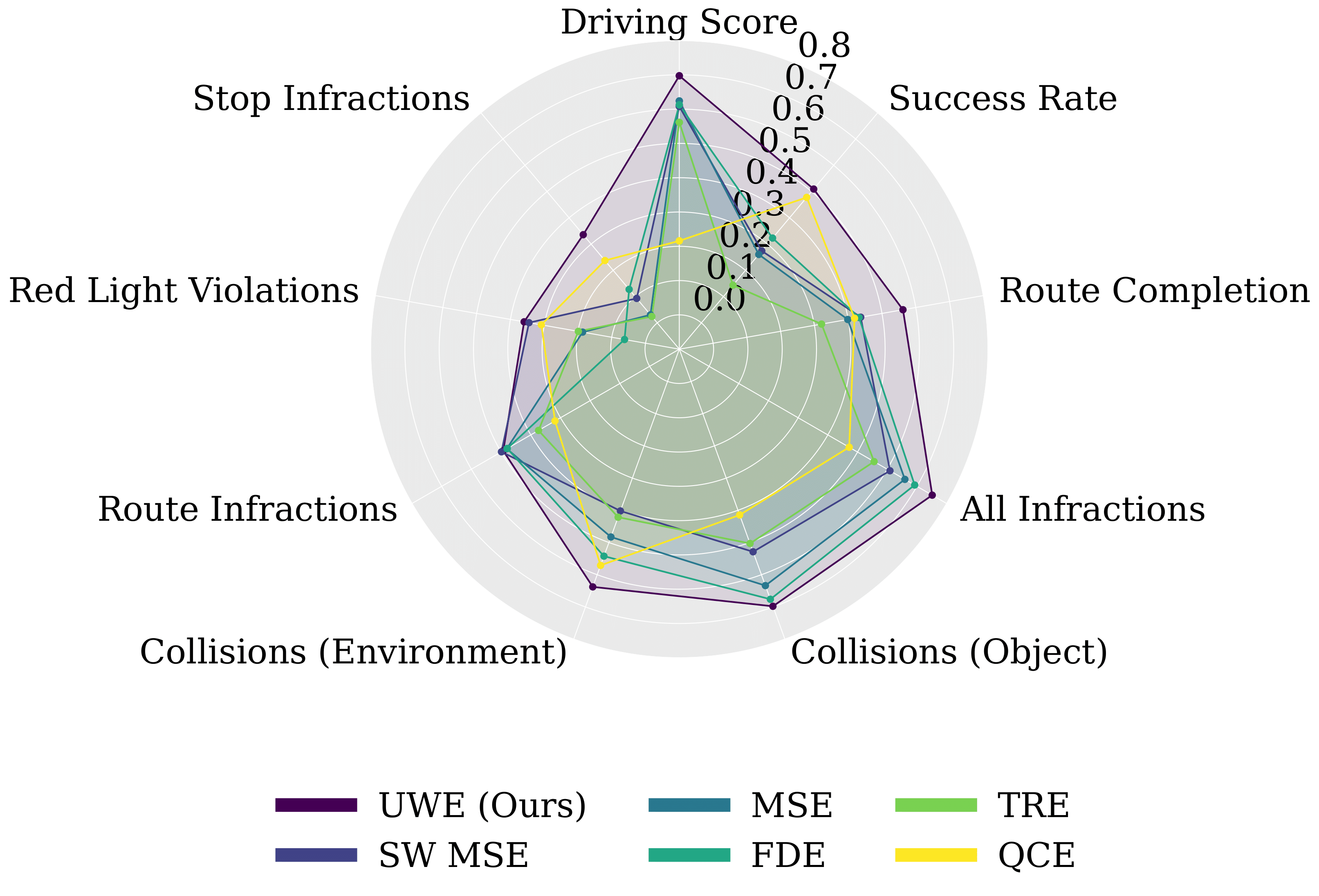}
    \vspace{-0.2cm}
    \caption{\textbf{Correlation Analysis in Simulation.} Best overall performance is shown by UWE (the proposed uncertainty-weighted error).}
    \label{fig:sim-steer-radar}
\end{figure}

\boldparagraph{Revisiting Correlation in Simulation}
As shown in Fig.~\ref{fig:ds-2x3}, we show an overall poor correlation (both for Driving Score and Success Rate, isolated in Fig.~\ref{fig:sim-steer-radar}) for standard offline metrics. Specifically, QCE and TRE, which are designed for higher correlation with the online Success Rate metric, show poor correlation (0.22 and 0.56, respectively) to more holistic online performance metrics such as the Driving Score. As shown in Fig.~\ref{fig:ds-2x3}, QCE and TRE even perform worse than FDE (waypoint-based final displacement error), which is a standard offline metric for high-performing systems today. We hypothesize that the reason these were successful is due to the highly simplistic test settings in Codevilla et al.~\cite{Codevilla2018ECCV} (Town02, not considering collisions or infractions). However, we observe that our proposed UWE achieves the highest correlation within this challenging benchmark, $0.69$, $0.08$ higher than prior offline metrics. We only show selected metrics, as the remaining analyzed offline metrics performed with worse correlation. 
We study nine online metrics to get a complete understanding of the strengths and weaknesses of these metrics and identify key aspects that help choose the most promising metric. Our UWE demonstrates a stronger correlation with all online metrics and significantly outperforms other common offline metric baselines, as shown in Fig.~\ref{fig:sim-steer-radar}. We also compare our UWE metric with the PDM (Predictive Driver Model) score from NAVSIM~\cite{dauner2024navsim}, as shown in Fig.~\ref{fig:ds-2x3}, using the publicly available code. As shown in Fig.~\ref{fig:ds-2x3}, the PDM score is lower in our more realistic CARLA simulation compared to its performance on nuPlan. Specifically, in settings with more complex physics and reactivity, UWE's correlation exhibits significantly higher fidelity than the PDM score. We note that this is likely because the PDM score makes assumptions about agent reactivity and environmental information (\ie, \textit{being overly accommodating}). In contrast, UWE does not require such assumptions. As such, our metric provides a simple yet effective and scalable alternative potentially applicable to a broader range of situations with various agent dynamics.

\subsection{Generalization to Real-World Settings}

We use two distinct real-world settings for evaluation: targeted and naturalistic driving settings. Physics in the real-world such as momentum, friction, and other dynamic-affecting constructs are simplified in simulation. Therefore, evaluations in simulation are not always indicative of the expectations of real-world evaluations.

\begin{figure}[!t]
    \centering
    \begin{tabular}{c}
       \includegraphics[width=0.46\linewidth]{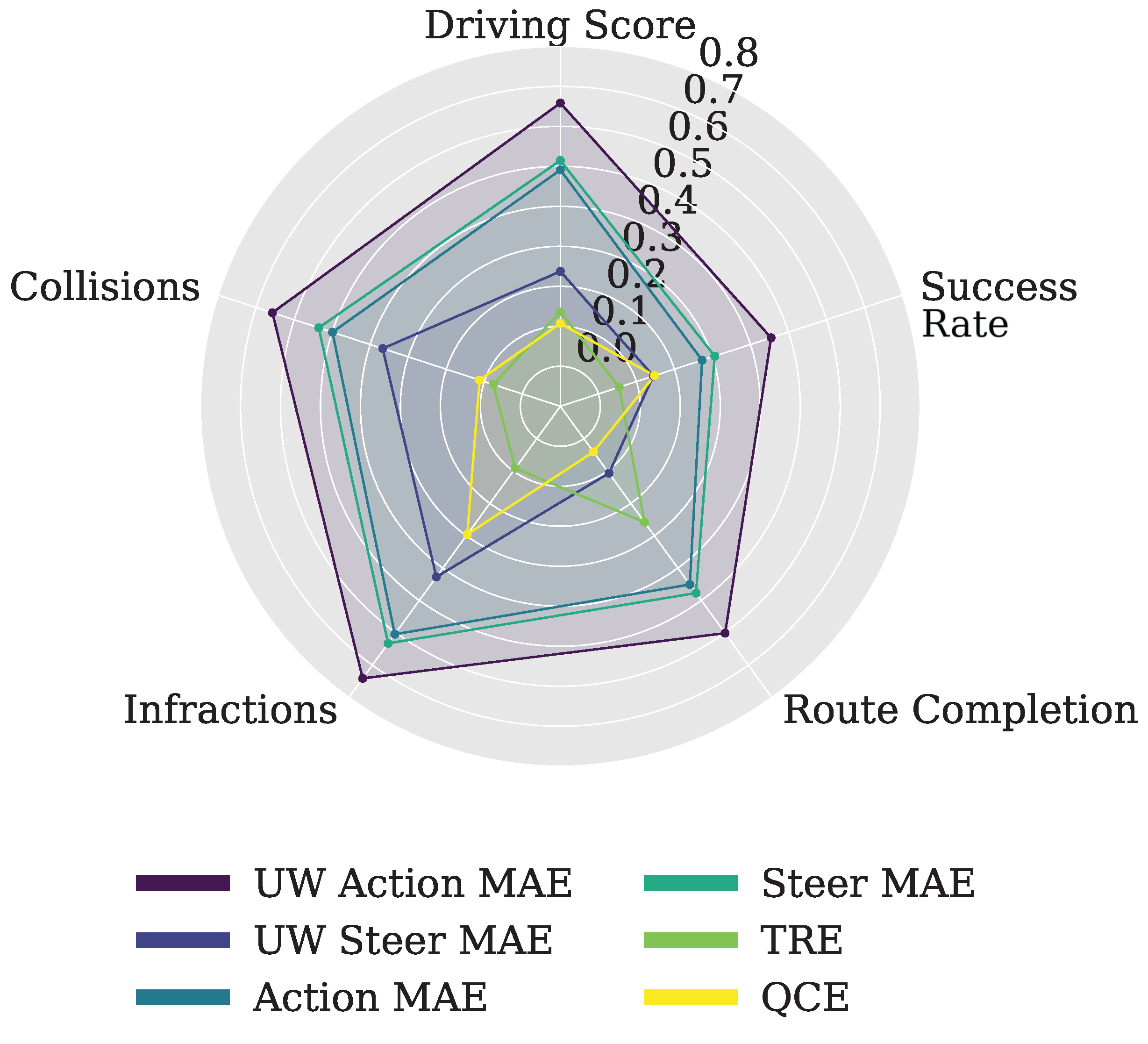}
       %& 
       \includegraphics[width=0.46\linewidth]{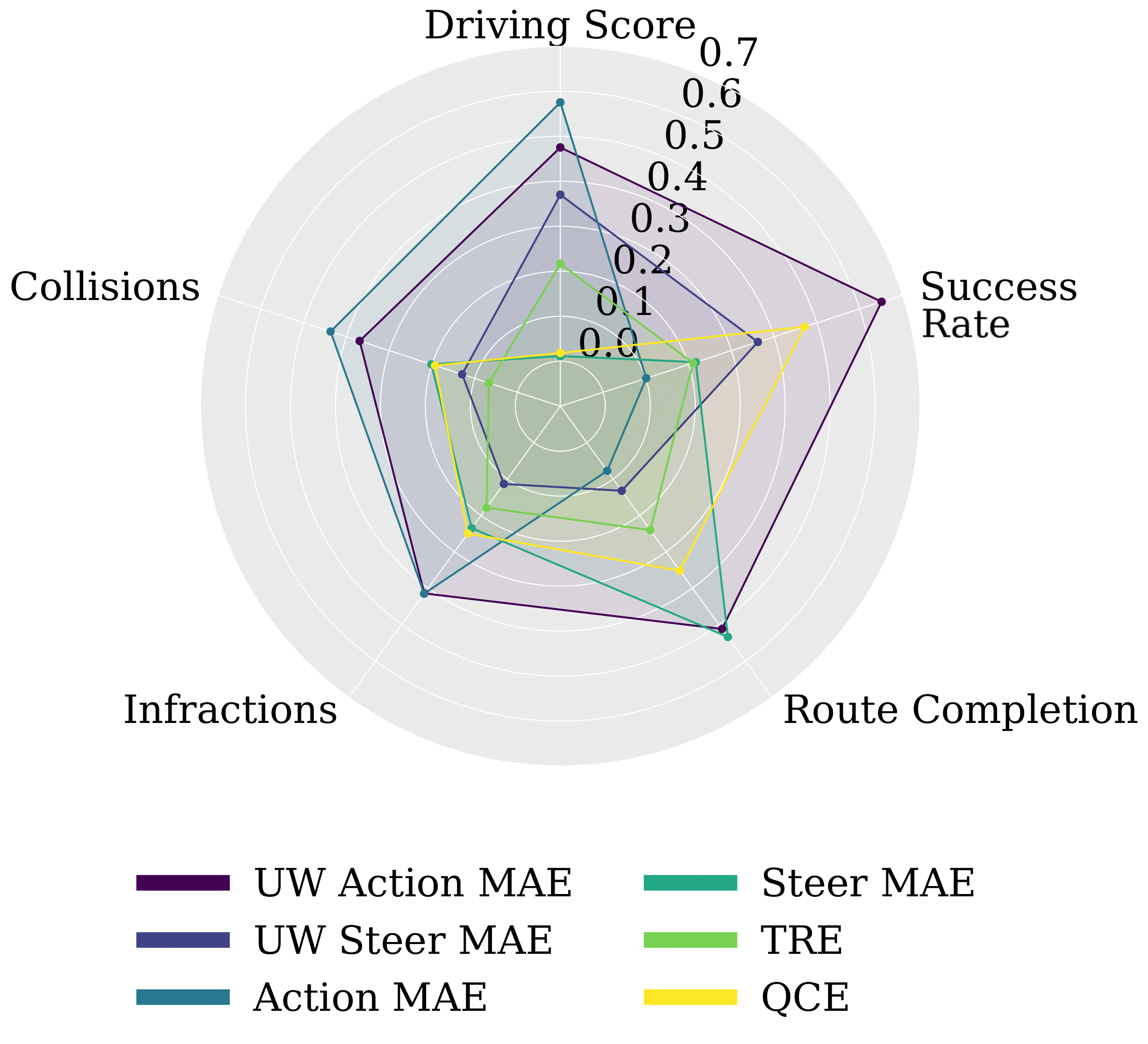}
    \end{tabular}
        \vspace{-0.2cm}
    \caption{\textbf{Correlation Analysis in the Real-World. } Results depict targeted scenarios evaluation (left) in the real-world (short routes over a traffic light or pedestrian scenario) and naturalistic longer routes (right). We find consistent trends with simulation-based results, \ie, UW action correlates well all online metrics when compared to the TRE, QCE, and baseline steer MAE and action MAE in targeted scenarios. Similarly, in longer routes, we show competitive correlation, \ie, compared to the baseline steer and action MAE, while outperforming TRE and QCE.
    }
    \label{fig:realworld-action-radar}
    \vspace{-0.5cm}
\end{figure}

\boldparagraph{Uncertainty Metrics in the Real-World}
Fig.~\ref{fig:realworld-action-radar} shows the effect of the UW action metric on the real-world evaluations. We compare our UW metric for both the full action or only steering against the baseline, as well as the standard TRE and QCE metrics. Due to complexity of computing some metrics in real-world conditions, we focus on the primary five online metrics. Consistent with our findings in the simulation, Fig.~\ref{fig:realworld-action-radar} shows that in our targeted scenario setting, the UW action metric has the highest correlation across all metrics in comparison to the baseline. This is further validated by the naturalistic driving setting where the UW action MAE shows a competitive correlation when compared to baseline steer-only or full action MAE for most of the online metrics.

\boldparagraph{Targeted Setting} We conduct experiments in a controlled targeted setting where the brightness and hue of the lighting are static and the floor is leveled. We deploy an Agile-X LIMO for data capturing and inference. Our targeted setting experiments include dense events such as stoplights and crossing automated mechatronic toys (robots, dinosaurs, etc). In Fig.~\ref{fig:realworld-action-radar}, our UW action MAE outperforms other metrics in all correlations with the online metrics.

\boldparagraph{Naturalistic Driving Setting} Uneven surfaces such as brick roads or potholes are common in naturalistic driving settings but are not yet simulated by CARLA. However, these have major impacts on driving performance in the real-world. Challenges of naturalistic driving settings include varying weather conditions and unexpected obstacles such as road debris. Hence, we mitigate a few of these challenges using hardware such as deploying a hacked suspension-based RC car and utilizing an automatic white-balancing Android phone camera. Additionally, we chose a pedestrian walkway plaza to run naturalistic driving evaluations as it is rich with obstacles, uneven surfaces, and diverse weather conditions. The complete setup is seen in Fig.~\ref{fig:rw-setup}. In Fig.~\ref{fig:realworld-action-radar}, we see that UW action MAE once again performs exceptionally well at indicating good online-driving performance, outperforming the other metrics when correlated with Success Rate and coming in second in correlation with Driving Score. We note that action MAE has a higher correlation to Driving Score than our UW action MAE, and that can be attributed to the greater variation between online and offline evaluation in naturalistic driving testing environments. 

\subsection{UWE Ablation}

We further investigate the reliability of the uncertainty estimates obtained via MC dropout. When we replace the dropout mechanism with a model ensemble and observe an increase in correlation from 0.69 to 0.80 in the Pearson coefficient. While this aligns with prior literature \cite{ovadia2019can}, dropout-based inference remains more efficient for larger models. To assess metric sensitivity, we recompute the correlation shown in Fig.~\ref{fig:ds-2x3}. For this test, we use uncertainty weights derived from a single model, keeping weights fixed rather than adapting them per model. Specifically, we compute the overall correlation with the DS using weights from a model trained for 10 epochs (achieving a 0.67 Pearson coefficient) and 40 epochs (Pearson coefficient of 0.73). These results indicate that UWE is not highly sensitive to model selection. While we evaluate UWE across models and deployment settings, including real-world applications on a physical platform, each metric introduces trade-offs. For instance, UWE does not require HD Maps or the meticulous offline simulation design used in NAVSIM \cite{dauner2024navsim}.
\section{Conclusion}
\label{sec:conclusion}
Offline evaluation offers several benefits for researchers and engineers, such as safety, scalability, and reduced cost. 
%, and efficient evaluation. of autonomous driving policies presents 
In this work, we seek to better characterize the correlation between offline and online evaluation. We do so through a series of experiments in both simulation and real-world. To ensure the broad impact of our findings, \ie, across diverse model types and conditions, we further incorporate diverse model sampling strategies in order. An uncertainty-weighted metric is shown to improve correlation with various online aspects of autonomous driving policies, \eg, over standard offline metrics used today. However, our study is a preliminary step toward efficient and scalable evaluation procedures. While our real-world platform (RC car) validates UWE’s feasibility, scaling to full-sized vehicles in complex urban environments remains future work. We pursue a finer-grained evaluation of various online metrics suitable for complex urban driving, predicting subtle errors can be difficult. There is a need to devise scalable and efficient metrics that can account for more subtle violations. We plan to study this in the future using our physical platform. Although we evaluate using a field study in the real-world, additional larger-scale evaluations, \eg, with full-scale vehicles and diverse controlled test routes, should be pursued to validate our framework and metrics further in the future. 

\boldparagraph{Acknowledgments}
We thank the Red Hat Collaboratory (award \#2024-01-RH07, \#2025-01-RH04) and the National Science Foundation (IIS-2152077) for supporting this research.

\bibliographystyle{IEEEtran}

\bibliography{egbib}

@String(PAMI = {IEEE Trans. Pattern Anal. Mach. Intell.})

@String(CVPR= {IEEE Conf. Comput. Vis. Pattern Recog.})

@String(ICCV= {Int. Conf. Comput. Vis.})

@String(ECCV= {Eur. Conf. Comput. Vis.})

@String(ICLR = {Int. Conf. Learn. Represent.})

@String(PAMI  = {IEEE TPAMI})

@String(CVPR  = {CVPR})

@String(ICCV  = {ICCV})

@String(ECCV  = {ECCV})

@String(ICLR  = {ICLR})

@inproceedings{gal2016dropout,
  title={Dropout as a bayesian approximation: Representing model uncertainty in deep learning},
  author={Gal, Yarin and Ghahramani, Zoubin},
  booktitle={ICML},
  year={2016},
}

@inproceedings{shao2023safety,
  title={Safety-enhanced autonomous driving using interpretable sensor fusion transformer},
  author={Shao, Hao and Wang, Letian and Chen, Ruobing and Li, Hongsheng and Liu, Yu},
  booktitle={CoRL},
  year={2023},
}

@inproceedings{hu2022st,
  title={{ST-P3}: End-to-end vision-based autonomous driving via spatial-temporal feature learning},
  author={Hu, Shengchao and Chen, Li and Wu, Penghao and Li, Hongyang and Yan, Junchi and Tao, Dacheng},
  booktitle={ECCV},
  year={2022},
}

@InProceedings{Cui_2021_ICCV,
    author    = {Cui, Alexander and Casas, Sergio and Sadat, Abbas and Liao, Renjie and Urtasun, Raquel},
    title     = {LookOut: Diverse Multi-Future Prediction and Planning for Self-Driving},
    booktitle = {ICCV},
    year      = {2021},
}

@inproceedings{zhang2022rethinking,
  title={Rethinking closed-loop training for autonomous driving},
  author={Zhang, Chris and Guo, Runsheng and Zeng, Wenyuan and Xiong, Yuwen and Dai, Binbin and Hu, Rui and Ren, Mengye and Urtasun, Raquel},
  booktitle={ECCV},
  year={2022},
}

@inproceedings{schreier2023offline,
  title={On Offline Evaluation of 3D Object Detection for Autonomous Driving},
  author={Schreier, Tim and Renz, Katrin and Geiger, Andreas and Chitta, Kashyap},
  booktitle={ICCV},
  year={2023}
}

@INPROCEEDINGS{chen2021what,
  author={Chen, Long and Platinsky, Lukas and Speichert, Stefanie and Osiński, Błażej and Scheel, Oliver and Ye, Yawei and Grimmett, Hugo and Del Pero, Luca and Ondruska, Peter},
  booktitle={ICRA}, 
  title={What data do we need for training an AV motion planner?}, 
  year={2021}
}

@article{gonzalez2015review,
  title={A review of motion planning techniques for automated vehicles},
  author={Gonz{\'a}lez, David and P{\'e}rez, Joshu{\'e} and Milan{\'e}s, Vicente and Nashashibi, Fawzi},
  journal={T-ITS},
  year={2015},
}

@inproceedings{xu2021autonomous,
  title={Autonomous vehicle motion planning via recurrent spline optimization},
  author={Xu, Wenda and Wang, Qian and Dolan, John M},
  booktitle={ICRA},
  year={2021},
}

@article{chitta2022transfuser,
  title={TransFuser: Imitation with Transformer-Based Sensor Fusion for Autonomous Driving},
  author={Chitta, Kashyap and Prakash, Aditya and Jaeger, Bernhard and Yu, Zehao and Renz, Katrin and Geiger, Andreas},
  journal={PAMI},
  year={2023}
}

@inproceedings{behl2020label,
  title={Label efficient visual abstractions for autonomous driving},
  author={Behl, Aseem and Chitta, Kashyap and Prakash, Aditya and Ohn-Bar, Eshed and Geiger, Andreas},
  booktitle={IROS},
  year={2020}
}

@inproceedings{prakash2021multi,
  title={Multi-modal fusion transformer for end-to-end autonomous driving},
  author={Prakash, Aditya and Chitta, Kashyap and Geiger, Andreas},
  booktitle={CVPR},
  year={2021}
}

@article{carlaleaderboard,
  title={Carla autonomous driving leaderboard},
  journal={https://leaderboard.carla.org/},
  year={2023}
}

@article{ovadia2019can,
  title={Can you trust your model's uncertainty? Evaluating predictive uncertainty under dataset shift},
  author={Ovadia, Yaniv and Fertig, Emily and Ren, Jie and Nado, Zachary and Sculley, David and Nowozin, Sebastian and Dillon, Joshua V and Lakshminarayanan, Balaji and Snoek, Jasper},
  journal={NeurIPS},
  year={2019}
}

@inproceedings{laskey2017dart,
  title={Dart: Noise injection for robust imitation learning},
  author={Laskey, Michael and Lee, Jonathan and Fox, Roy and Dragan, Anca and Goldberg, Ken},
  booktitle={CoRL},
  year={2017}
}

@article{jiang2023vad,
  title={{VAD}: Vectorized Scene Representation for Efficient Autonomous Driving},
  author={Jiang, Bo and Chen, Shaoyu and Xu, Qing and Liao, Bencheng and Chen, Jiajie and Zhou, Helong and Zhang, Qian and Liu, Wenyu and Huang, Chang and Wang, Xinggang},
  journal={ICCV},
  year={2023}
}

@inproceedings{li2020end,
  title={End-to-end contextual perception and prediction with interaction transformer},
  author={Li, Lingyun Luke and Yang, Bin and Liang, Ming and Zeng, Wenyuan and Ren, Mengye and Segal, Sean and Urtasun, Raquel},
  booktitle={IROS},
  year={2020},
}

@inproceedings{filos2020can,
  title={Can autonomous vehicles identify, recover from, and adapt to distribution shifts?},
  author={Filos, Angelos and Tigas, Panagiotis and McAllister, Rowan and Rhinehart, Nicholas and Levine, Sergey and Gal, Yarin},
  booktitle={ICML},
  year={2020}
}

@article{bojarski2016end,
  title={End to end learning for self-driving cars},
  author={Bojarski, Mariusz and Del Testa, Davide and Dworakowski, Daniel and Firner, Bernhard and Flepp, Beat and Goyal, Prasoon and Jackel, Lawrence D and Monfort, Mathew and Muller, Urs and Zhang, Jiakai and others},
  journal={arXiv preprint arXiv:1604.07316},
  year={2016}
}

@inproceedings{wang2019monocular,
  title={Monocular plan view networks for autonomous driving},
  author={Wang, Dequan and Devin, Coline and Cai, Qi-Zhi and Kr{\"a}henb{\"u}hl, Philipp and Darrell, Trevor},
  booktitle={IROS},
  year={2019}
}

@inproceedings{kendall2019learning,
  title={Learning to drive in a day},
  author={Kendall, Alex and Hawke, Jeffrey and Janz, David and Mazur, Przemyslaw and Reda, Daniele and Allen, John-Mark and Lam, Vinh-Dieu and Bewley, Alex and Shah, Amar},
  booktitle={ICRA},
  year={2019}
}

@inproceedings{xu2017end,
  title={End-to-end learning of driving models from large-scale video datasets},
  author={Xu, Huazhe and Gao, Yang and Yu, Fisher and Darrell, Trevor},
  booktitle={CVPR},
  year={2017}
}

@inproceedings{hawke2020urban,
  title={Urban driving with conditional imitation learning},
  author={Hawke, Jeffrey and Shen, Richard and Gurau, Corina and Sharma, Siddharth and Reda, Daniele and Nikolov, Nikolay and Mazur, Przemys{\l}aw and Micklethwaite, Sean and Griffiths, Nicolas and Shah, Amar and others},
  booktitle={ICRA},
  year={2020},
}

@article{henaff2019model,
  title={Model-predictive policy learning with uncertainty regularization for driving in dense traffic},
  author={Henaff, Mikael and Canziani, Alfredo and LeCun, Yann},
  journal={ICLR},
  year={2019}
}

@inproceedings{prakash2020exploring,
  title={Exploring data aggregation in policy learning for vision-based urban autonomous driving},
  author={Prakash, Aditya and Behl, Aseem and Ohn-Bar, Eshed and Chitta, Kashyap and Geiger, Andreas},
  booktitle={CVPR},
  year={2020}
}

@InProceedings{Dosovitskiy2017CORL,
  author    = {Alexey Dosovitskiy and German Ros and Felipe Codevilla and Antonio Lopez and Vladlen Koltun},
  title     = {{CARLA}: {An} Open Urban Driving Simulator},
  booktitle = {CoRL},
  year      = {2017},
}

@inproceedings{Codevilla2018ICRA,
  author    = {Felipe Codevilla and
               Matthias Miiller and
               Antonio L{\'{o}}pez and
               Vladlen Koltun and
               Alexey Dosovitskiy},
  title     = {End-to-End Driving Via Conditional Imitation Learning},
  booktitle = {ICRA},
  year      = {2018},
}

@inproceedings{lbc,
  title={Learning by cheating},
  author={Chen, Dian and Zhou, Brady and Koltun, Vladlen and Kr{\"a}henb{\"u}hl, Philipp},
  booktitle={CoRL},
  year={2020},
}

@inproceedings{bansal2018chauffeurnet,
  author    = {Bansal, Mayank and Krizhevsky, Alex and Ogale, Abhijit},
  title     = {{ChauffeurNet}: Learning to drive by imitating the best and synthesizing the worst},
  booktitle = {RSS},
  year      = {2019},
}

@inproceedings{Codevilla2018ECCV,
  author    = {Codevilla, Felipe and Lopez, Antonio M. and Koltun, Vladlen and Dosovitskiy, Alexey},
  title     = {On Offline Evaluation of Vision-based Driving Models},
  booktitle = {ECCV},
  year      = {2018},
}

@article{osa2018algorithmic,
  title={An algorithmic perspective on imitation learning},
  author={Osa, Takayuki and Pajarinen, Joni and Neumann, Gerhard and Bagnell, J Andrew and Abbeel, Pieter and Peters, Jan and others},
  journal={Foundations and Trends{\textregistered} in Robotics},
  year={2018},
}

@inproceedings{ross2011reduction,
  title={A reduction of imitation learning and structured prediction to no-regret online learning},
  author={Ross, St{\'e}phane and Gordon, Geoffrey and Bagnell, Drew},
  booktitle={AISTATS},
  year={2011}
}

@inproceedings{muller2018driving,
  author    = {Matthias M{\"{u}}ller and
               Alexey Dosovitskiy and
               Bernard Ghanem and
               Vladlen Koltun},
  title     = {Driving Policy Transfer via Modularity and Abstraction},
  booktitle = {CoRL},
  year      = {2018},
}

@inproceedings{pomerleau1989alvinn,
  title={{ALVINN}: An autonomous land vehicle in a neural network},
  author={Pomerleau, Dean A},
  booktitle={NeurIPS},
  year={1989}
}

@inproceedings{hu2021safe,
  title={Safe Local Motion Planning with Self-Supervised Freespace Forecasting},
  author={Hu, Peiyun and Huang, Aaron and Dolan, John and Held, David and Ramanan, Deva},
  booktitle={CVPR},
  year={2021}
}

@article{hanselmann2022king,
  title={KING: Generating Safety-Critical Driving Scenarios for Robust Imitation via Kinematics Gradients},
  author={Hanselmann, Niklas and Renz, Katrin and Chitta, Kashyap and Bhattacharyya, Apratim and Geiger, Andreas},
  journal={ECCV},
  year={2022}
}

@inproceedings{caesar2020nuscenes,
  title={nuscenes: A multimodal dataset for autonomous driving},
  author={Caesar, Holger and Bankiti, Varun and Lang, Alex H and Vora, Sourabh and Liong, Venice Erin and Xu, Qiang and Krishnan, Anush and Pan, Yu and Baldan, Giancarlo and Beijbom, Oscar},
  booktitle={CVPR},
  year={2020}
}

@inproceedings{wang2021uncertainty,
  title={Uncertainty-Aware Pseudo Label Refinery for Domain Adaptive Semantic Segmentation},
  author={Wang, Yuxi and Peng, Junran and Zhang, ZhaoXiang},
  booktitle={ICCV},
  year={2021}
}

@inproceedings{ohn2020learning,
  title={Learning Situational Driving},
  author={Ohn-Bar, Eshed and Prakash, Aditya and Behl, Aseem and Chitta, Kashyap and Geiger, Andreas},
  booktitle={CVPR},
  year={2020}
}

@inproceedings{lai2025zerovo,
  title={{ZeroVO}: Visual Odometry with Minimal Assumptions},
  author={Lai, Lei and Yin, Zekai and Ohn-Bar, Eshed},
  booktitle={CVPR},
  year={2025}
}

@inproceedings{lai2024uncertainty,
  title={Uncertainty-Guided Never-Ending Learning to Drive},
  author={Lai, Lei and Ohn-Bar, Eshed and Arora, Sanjay and Yi, John Seon Keun},
  booktitle={CVPR},
  year={2024}
}

@inproceedings{zhang2023coaching,
  title={Coaching a Teachable Student},
  author={Zhang, Jimuyang and Huang, Zanming and Ohn-Bar, Eshed},
  booktitle={CVPR},
  year={2023}
}

@article{zhu2023learning,
  title={Learning to drive anywhere},
  author={Zhu, Ruizhao and Huang, Peng and Ohn-Bar, Eshed and Saligrama, Venkatesh},
  journal={CoRL},
  year={2023}
}

@inproceedings{lai2023xvo,
  title={{XVO}: Generalized Visual Odometry via Cross-Modal Self-Training},
  author={Lai, Lei and Shangguan, Zhongkai and Zhang, Jimuyang and Ohn-Bar, Eshed},
  booktitle={ICCV},
  year={2023} 
}

@inproceedings{zhang2024feedback,
  title={Feedback-Guided Autonomous Driving},
  author={Zhang, Jimuyang and Huang, Zanming and Ray, Arijit and Ohn-Bar, Eshed},
  booktitle={CVPR},
  year={2024}
}

@inproceedings{kendall2018multi,
  title={Multi-task learning using uncertainty to weigh losses for scene geometry and semantics},
  author={Kendall, Alex and Gal, Yarin and Cipolla, Roberto},
  booktitle={CVPR},
  year={2018}
}

@INPROCEEDINGS{tang2018teaching,
  author={Tang, Jie and Shaoshan, Liu and Pei, Songwen and Zuckerman, Stéphane and Chen, Liu and Shi, Weisong and Gaudiot, Jean-Luc},
  booktitle={COMPSAC}, 
  title={Teaching Autonomous Driving Using a Modular and Integrated Approach}, 
  year={2018}
}

@article{jaeger2023hidden,
  title={Hidden Biases of End-to-End Driving Models},
  author={Jaeger, Bernhard and Chitta, Kashyap and Geiger, Andreas},
  journal={ICCV},
  year={2023}
}

@inproceedings{hu2023planning,
  title={Planning-oriented autonomous driving},
  author={Hu, Yihan and Yang, Jiazhi and Chen, Li and Li, Keyu and Sima, Chonghao and Zhu, Xizhou and Chai, Siqi and Du, Senyao and Lin, Tianwei and Wang, Wenhai and others},
  booktitle={CVPR},
  year={2023}
}

@article{dauner2023parting,
  title={Parting with Misconceptions about Learning-based Vehicle Motion Planning},
  author={Dauner, Daniel and Hallgarten, Marcel and Geiger, Andreas and Chitta, Kashyap},
  journal={arXiv preprint arXiv:2306.07962},
  year={2023}
}

@article{li2023ego,
  title={Is Ego Status All You Need for Open-Loop End-to-End Autonomous Driving?},
  author={Li, Zhiqi and Yu, Zhiding and Lan, Shiyi and Li, Jiahan and Kautz, Jan and Lu, Tong and Alvarez, Jose M},
  journal={arXiv preprint arXiv:2312.03031},
  year={2023}
}

@inproceedings{wang2021advsim,
  title={Advsim: Generating safety-critical scenarios for self-driving vehicles},
  author={Wang, Jingkang and Pun, Ava and Tu, James and Manivasagam, Sivabalan and Sadat, Abbas and Casas, Sergio and Ren, Mengye and Urtasun, Raquel},
  booktitle={CVPR},
  year={2021}
}

@inproceedings{tu2023towards,
  title={Towards Scalable Coverage-Based Testing of Autonomous Vehicles},
  author={Tu, James and Suo, Simon and Zhang, Chris and Wong, Kelvin and Urtasun, Raquel},
  booktitle={CoRL},
  year={2023}
}

@article{weber2019framework,
  title={A framework for definition of logical scenarios for safety assurance of automated driving},
  author={Weber, Hendrik and Bock, Julian and Klimke, Jens and Roesener, Christian and Hiller, Johannes and Krajewski, Robert and Zlocki, Adrian and Eckstein, Lutz},
  journal={Traffic Injury Prevention},
  year={2019}
}

@article{o2018scalable,
  title={Scalable end-to-end autonomous vehicle testing via rare-event simulation},
  author={O'Kelly, Matthew and Sinha, Aman and Namkoong, Hongseok and Tedrake, Russ and Duchi, John C},
  journal={NeurIPS},
  year={2018}
}

@inproceedings{menzel2018scenarios,
  title={Scenarios for development, test and validation of automated vehicles},
  author={Menzel, Till and Bagschik, Gerrit and Maurer, Markus},
  booktitle={IV},
  year={2018}
}

@article{lee2020adaptive,
  title={Adaptive stress testing: Finding likely failure events with reinforcement learning},
  author={Lee, Ritchie and Mengshoel, Ole J and Saksena, Anshu and Gardner, Ryan W and Genin, Daniel and Silbermann, Joshua and Owen, Michael and Kochenderfer, Mykel J},
  journal={Journal of Artificial Intelligence Research},
  year={2020}
}

@techreport{thorn2018framework,
  title={A framework for automated driving system testable cases and scenarios},
  author={Thorn, Eric and Kimmel, Shawn C and Chaka, Michelle and Hamilton, Booz Allen and others},
  year={2018},
  institution={Department of Transportation}
}

@misc{uber,
  title = {{The New York Times}. {S}elf-Driving Uber Car Kills Pedestrian in Arizona, Where Robots Roam},
  publisher = {The New York Times},
  year = {2018},
}

@misc{cruiseend,
  author =       {Government-Fleet},
  title =        {California {DMV} Removes Cruise's Driverless Vehicle Testing Permits},
  year =         {2023}
}

@misc{waymodrunk,
  author =       {{The Last Driver License Holder}},
  title =        {\href{{https://thelastdriverlicenseholder.com/2024/05/14/waymo-confused-behind-a-trailer-with-a-tree/}}{      Waymo Confused Behind a Trailer with a Tree}},
  year =         {2024}
}

@misc{xmax,
  title = {{Traxxas X-Maxx RC Vehicle}, \url{https://traxxas.com/products/landing/x-maxx/}},
}

@article{ghorai2024causation,
  title={A Causation Analysis of Autonomous Vehicle Crashes},
  author={Ghorai, P. and others},
  journal={ITSM},
  year={2024},
}

@article{wu2023policy,
  title={Policy pre-training for end-to-end autonomous driving via self-supervised geometric modeling},
  author={Wu, Penghao and Chen, Li and Li, Hongyang and Jia, Xiaosong and Yan, Junchi and Qiao, Yu},
  journal={ICLR},
  year={2023}
}

@inproceedings{liu2022convnet,
  title={A convnet for the 2020s},
  author={Liu, Zhuang and Mao, Hanzi and Wu, Chao-Yuan and Feichtenhofer, Christoph and Darrell, Trevor and Xie, Saining},
  booktitle={CVPR},
  year={2022}
}

@inproceedings{radosavovic2020designing,
  title={Designing network design spaces},
  author={Radosavovic, Ilija and Kosaraju, Raj Prateek and Girshick, Ross and He, Kaiming and Doll{\'a}r, Piotr},
  booktitle={CVPR},
  year={2020}
}

@article{stachowicz2024racer,
  title={{RACER}: Epistemic Risk-Sensitive RL Enables Fast Driving with Fewer Crashes},
  author={Stachowicz, Kyle and Levine, Sergey},
  journal={arXiv preprint arXiv:2405.04714},
  year={2024}
}

@article{off-RL-tutorial,
  title={Offline reinforcement learning: Tutorial, review, and perspectives on open problems},
  author={Sergey Levine and Aviral Kumar and George Tucker and Justin Fu},
  year={2020},
  journal={arXiv preprint arXiv:2005.01643}
}

@article{review-off-RL-tutorial,
  title={A review of off-policy evaluation in reinforcement learning},
  author={Uehara, Shi, Kallus},
  year={2022},
  journal={arXiv preprint arXiv:2212.06355}
}

@inproceedings{instantiation-uncertainty-based-metric,
  title={Deeply-debiased off-policy interval estimation},
  author={Chengchun Shi and Runzhe Wan and Victor Chernozhukov and Rui Song},
  booktitle={international conference on machine learning},
  year={2021},
}

@inproceedings{dauner2024navsim,
title={{NAVSIM}: Data-Driven Non-Reactive Autonomous Vehicle Simulation and Benchmarking},
author={Daniel Dauner and Marcel Hallgarten and Tianyu Li and Xinshuo Weng and Zhiyu Huang and Zetong Yang and Hongyang Li and Igor Gilitschenski and Boris Ivanovic and Marco Pavone and Andreas Geiger and Kashyap Chitta},
booktitle={NeurIPS},
year={2024}
}

\end{document}